# A Comparison of Deep Learning Models for the Prediction of Hand Hygiene Videos

**[1]Rashmi Bakshi**

[1]Assistant Lecturer, part-time, School of Electrical and Electronic Engineering, TU Dublin, City campus

Email: [1]rashmi.bakshi@tudublin.ie
Contact: [1]0899885491

**Abstract:** This paper presents a comparison of various deep learning models such as Exception, Resnet-50 and Inception V3 for the classification and prediction of hand hygiene gestures, which were recorded in accordance with the World Health Organization (WHO) guidelines. The dataset consists of six hand hygiene movements in a video format, gathered for 30 participants. The network consists pre-trained models with image net weights and a modified head of the model. An accuracy of 37% (Xception model), 33% (Inception V3) and 72% (ResNet-50) is achieved in the classification report after the training of the models for 25 epochs. ResNet-50 model clearly outperforms with correct class predictions. The major speed limitation can be overcome with the use of fast processing GPU for future work. A complete hand hygiene dataset along with other generic gestures such as one-hand movements (linear hand motion; circular hand rotation) will be tested with ResNet-50 architecture and the variants for health care workers.

*Index terms:* Deep Learning, Transfer learning, ImageNet, Hand Hygiene, Hand washing

## I. INTRODUCTION

Hospital Acquired Infections (HAIs) have a significant impact on the quality of life and result in an increase in the health care expenditure. According to the European Centre for Disease Prevention and Control (ECDC), 2.5 million cases of HAIs occur in European Union and European Economic Area (EU/EAA) each year, corresponding to 2.5 million DALYs (Disability Adjusted Life Year) which is a measure of the number of years lost due to ill health, disability or an early death [1]. MRSA-Methicillin Resistant Staphylococcus Aureus is a common bacteria associated with the spread of HAIs [2]. One method to prevent the cross transmission of these microorganisms is the implementation of well-structured hand hygiene practices. The World Health Organization (WHO) has provided guidelines about hand washing procedures for health care workers [3]. Best hand hygiene practices have been proven to reduce the rate of MRSA infections in a health care setting [4]. One challenge in dynamic healthcare environments is to ensure compliance with these hand hygiene guidelines and to evaluate the quality of hand washing. This is often done through auditing involving human observation. The hand washing process, however, is well structured and has particular dynamic hand gestures associated with each hand washing stage. One potential approach is to use imaging techniques to detect fine hand movements and identify user gestures, provide feedback to the user or a central management system, with the overall goal being an automated tool that can ensure compliance with the hand washing guidelines. In advance of developing hand hygiene stage detection system primarily for health care workers, preliminary results for distinct hand hygiene gestures are discussed in this work.

The aim of this paper is to explore the field of deep learning/ transfer learning for the purpose of classification of various hand hygiene stages. Keras models such as Xception, Resnet-50 and Inception V3 are utilized for the classification of hand hygiene video recordings with a future goal to detect hand hygiene stages in real time for health care workers.

## II. RELATED WORK

Deep learning is an emerging approach and has been widely applied in traditional artificial intelligence domains such as semantic parsing, transfer learning, computer vision, natural language processing and more [5]. Over the years, deep learning has gained increasing attention due to the significant low cost of computing hardware and access to high processing power (eg-GPU units) [5]. Deep learning solutions have shown promising results in applications such as text recognition [7], sound prediction [8], and image annotation [6]. Deep learning models essentially require thousands of data samples and heavy computational resources such as GPU for accurate classification and prediction analysis. However, there is a branch of machine learning, popularly known as "transfer learning" that does not necessarily requires large amounts of data for evaluation. Transfer learning is a machine learning technique wherein a model developed for one task is reused for the second related task. It refers to the situation where "finding" of one setting is exploited to improve the optimization in another setting [9]. Transfer learning is usually applied to the new dataset, which is generally smaller than the original data set, used to train the pre-trained model. In this work, pre-trained models on ImageNet dataset are applied for the classification of hand hygiene gestures.

An aluminium rig was constructed in order to record hand hygiene gestures for 30 participants. The data format was mp4 file that consists of six hand hygiene movements with a video length of 25-30 seconds. The video recordings were segmented into frames/images and six classes with labelled data was created [10]. A subset of this dataset is used in this work. Ivanovs et al. [11] train the neural network on labelled hand washing dataset captured in a health care setting , apply pre trained neural network models such as MobileNetV2 and Xception with >64% accuracy in recognizing hand washing gestures. In our previous work, Resnet-50 model was applied to hand hygiene dataset (set1 -25 epochs and set2-50 epochs) with an accuracy of 44% and 72% respectively. [12]

Keras-Deep learning models

High-level library, Keras [13] provides an open source implementation of various deep learning models that are made available alongside pre-trained weights. These models can be further used for prediction, feature extraction and fine-tuning. Models such as Xception, Resnet-50 and Inception V3 are applied for hand-hygiene gesture prediction in this paper.

1) Xception Model

The Xception architecture has 36 convolutional layers forming the feature extraction base of the network. The 36 convolutional layers are divided into 14 modules that have linear residual connections around them, except for the first and last modules. The Xception architecture is a linear stack of depth wise separable convolution layers with residual connections thereby making the architecture very easy to define and modify [14].

The model outperforms image net results for Resnet-50 and Inception V3. It can be seen in Table 3.

2) ResNet-50 Model

The ResNet architecture is based on the residual network where the problem of degradation is addressed by a deep residual learning framework in which stacked layers fit a residual mapping instead of the original mapping. ResNet50 is 50 layer network, a variant based on original 34 layer ResNet where each 2 layer block in 34 layer network is replaced by 3-layer bottleneck blocks to improve the accuracy and reduce the training time ; ImageNet database [15].

3) GoogleNet Inception V3

The network consists of 316 layers including convolution, average pool, max pool, concatenation, dropout, fully connection, and soft max layer. Factorization is an important feature of Inception v3 architecture, which factorizes a big kernel into smaller ones, resulting in less parameters and higher speed during the training process and avoids overfitting [16]. Table 3 provides the comparison of three models where top-1 and top-5 accuracy refers to the model's performance on the ImageNet validation dataset. Depth refers to the topological depth of the network that includes activation layers, batch normalization layers. Computational time is per inference step and an average of 30 batches and 10 repetitions [18].

III. METHODOLOGY

A subset of robust hand hygiene dataset is utilised with three hand hygiene classes and >1000 images in each class. The data is equally distributed among all the classes in order to avoid the bias during the training of the network. Table 2 lists the class label and number of images associated with each class. The hardware used for all the experiments were Intel(R) Core(TM) i5-5200U CPU @ 2.20GHz 2.19GHz with 4 GB RAM with 64-bit Windows Operating System.

In this work, two python scripts were created. Train.py is used for building the network with pre-trained model on ImageNet database with the help of an open source Keras library [13]. The head of the model was replaced with a new set of fully connected layers with random initializations. All layers below the head were frozen so that their weights cannot be updated. 'layer.trainable = False'. The model implementation is adapted from [17] where the author has implemented a multi classification system for 'sports' related video recordings and has achieved an accuracy higher than 90%.

*Algorithm and workflow for python script, train.py*

1. Read all the images in the dataset. Add labels to the images.
2. Convert the data to numpy arrays.
3. Perform one-hot encoding on the labels.
4. Split the data into training and testing set(75% training; 25% test)
5. Initialise the data augmentation for training and testing/validation set.
6. Load the 'Xception/ResNet/InceptionV3' model. Ensure the head of FC- fully connected layer is off. Weights-'imagenet'; include_top=False
7. Construct the head of the model to be placed on the top of the base model
8. Loop over all the layers in base model and freeze them so they will not be updated during the training process. Layer.trainable=False.
9. Compile the model with SGD optimizer.
10. Train the network
11. Print the classification report and loss-accuracy plot. Print the confusion matrix with accuracy, sensitivity and specificity.
12. Save the model

Predict.py was created to predict the class label for hand hygiene video recording that the network has not seen before.

*Workflow for python script, prediction.py*

1. Load the saved model and the label binarizer.
2. Initialise the image mean for mean subtraction along with the prediction queue.
3. Initialise the video stream, pointer to the output video file and frame dimensions; resize
4. .Loop over frames from the video file and make predictions on each frame. Update the prediction queue.
5. Write the output frame to the disk.

| Class Label | Number of Images |
|---|---|
| Fingers Interlaced | 1,346 |
| Fingers Interlocked | 1,349 |
| Palm2Palm | 1,329 |

Table 1. Class label and no. of images

## IV.     RESULTS

Three pre-trained models; Xception, Resnet-50 and Inception V3 were trained with a subset of hand hygiene images, which were captured for 30 participants while performing hand hygiene movements as per the World Health Organization guidelines. Twenty-Five training steps were selected for each model. Inception V3 was the fastest with less training time taken but the accuracy level achieved is 33%. ResNet-50 model outperforms Xception and Inception model with 72% accuracy. Speed-accuracy trade off can be overcome with the use of heavy computational resources-GPU for the future work.

Table 2 Classification Report for Xception-25 epochs

| Class Label | Precision | Recall | F1-score | Support |
|---|---|---|---|---|
| Fingers Interlaced | 0.45 | 0.10 | 0.16 | 337 |
| Fingers Interlocked | 0.35 | 0.72 | 0.47 | 337 |
| Palm2Palm | 0.42 | 0.30 | 0.35 | 332 |
| Micro average | **0.37** | **0.37** | **0.37** | **1006** |
| Macro average | 0.40 | 0.37 | 0.33 | 1006 |
| Weighted average | **0.40** | **0.37** | 0.33 | 1006 |

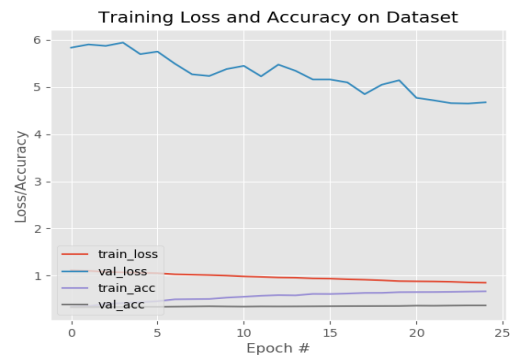

**Figure 1. Loss/accuracy curve for Xception model.**

| Model | Size (MB) | Top-1 Accuracy | Top-5 Accuracy | Depth | Time (ms) CPU |
|---|---|---|---|---|---|
| Xception | 88 | 0.790 | 0.945 | 126 | 109.42 |
| ResNet 50 | 98 | 0.749 | 0.921 | - | 58.20 |
| Inception V3 | 92 | 0.779 | 0.937 | 159 | 42.25 |

Table 3 A comparison of Xception, ResNet50 and Inceptionv3 from Keras Documentation [18]

Table 4: Classification Report for Inception V3, 25 epochs

| Class Label | Precision | Recall | F1-score | Support |
|---|---|---|---|---|
| Fingers Interlaced | 0.00 | 0.00 | 0.00 | 337 |
| Fingers Interlocked | 0.33 | 0.99 | 0.50 | 337 |
| Palm2Palm | 0.00 | 0.00 | 0.00 | 332 |
| Micro average | **0.33** | **0.33** | **0.33** | **1006** |
| Macro average | 0.11 | 0.33 | 0.17 | 1006 |
| Weighted average | **0.11** | **0.33** | 0.17 | 1006 |

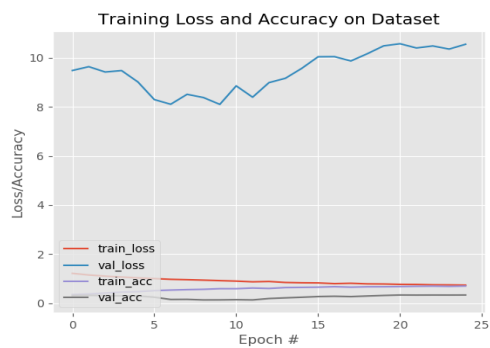

**Figure 2. Loss/Accuracy plot for Inception V3**

Table 5. Classification Report for ResNet-50, 25 epochs.

| Class Label | Precision | Recall | F1-score | Support |
|---|---|---|---|---|
| Fingers Interlaced | 0.58 | 0.95 | 0.72 | 337 |
| Fingers Interlocked | 0.89 | 0.92 | 0.91 | 337 |
| Palm2Palm | 0.93 | 0.29 | 0.44 | 332 |
| Micro average | **0.72** | **0.72** | **0.72** | **1006** |
| Macro average | 0.80 | 0.72 | 0.69 | 1006 |
| Weighted average | **0.80** | **0.72** | 0.69 | 1006 |

Table 2 is the classification report produced as the result of training the network with Xception Model. Figure 1 is the loss-accuracy curve for training and validation set and can be clearly seen that the loss is >0 and accuracy is <1.Table 4 is the classification report with precision, recall and F1 score for InceptionV3 model and Figure 2 is the loss-accuracy plot for InceptionV3 model.

Table 5 is the classification report for ResNet50 model with 72% accuracy as micro and 80% accuracy as weighted. It is evident with Figure 3 that ResNet50 makes correct class predictions for the unknown video frames in comparison to Xception, and Inception model and will be applied to a complete hand hygiene dataset in future.

TOP

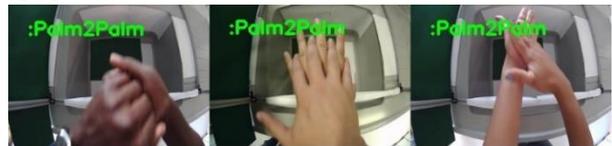

MIDDLE

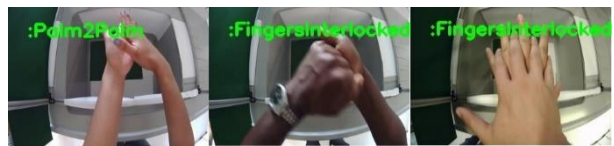

BOTTOM

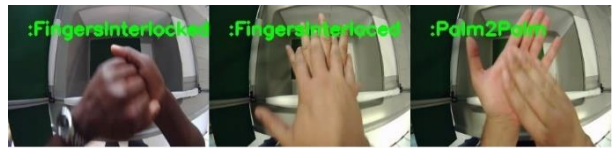

**Figure 3: Selected video frames prediction TOP (Xception) MIDDLE (Inception V3), BOTTOM (Resnet-50)**

*Supplementary Materials*: The complete robust hand hygiene dataset recorded for 30 participants along with the python code, model and results for the following are available online:

https://tudublin-my.sharepoint.com/personal/d16126930_mytudublin_ie/_layouts/15/onedrive.aspx?csf=1&web=1&e=mMwzfp%2E&cid=c37eef1d%2D41a7%2D4f36%2D8e8a%2Dd4c7ab03282c&FolderCTID=0x012000490722E83CFA9B4093B6BD80E02E3C07&id=%2Fpersonal%2Fd16126930%5Fmytudublin%5Fie%2FDocuments%2FHand%20Hygiene%20Research

## REFERENCES

[1] A. Cassini et al.Burden of Six Healthcare-Associated Infections on European Population Health: Estimating Incidence-Based Disability-Adjusted Life Years through a Population Prevalence-Based Modelling Study. PLoS Med., 2016,vol. 13, no. 10, pp. 1–16

[2] M. ŠLŠirak; A. Zvizdić., ; M. Hukić, Methicillin-resistant staphylococcus aureus (MRSA) as a cause of nosocomial wound infections. Bosn. J. Basic Med. Sci., 2010, vol. 10, no. 1, pp. 32–37.

[3] WHO Guidelines on Hand Hygiene in Healthcare. Available online: URL,http://apps.who.int/iris/bitstream/handle/10665/44102/9789241597906_eng.pdf;jsessionid=5FCA460BF88296 1840326820865681E8?sequence=1 (accessed 19-09-2021)

[4] Sroka S, Gastmerer P, Meyer E. Impact of alcohol hand-rub use on methicillin-resistant Staphylococcus aureus: an analysis of the literature. J Hosp Infect.2010,4(3): 204-11

[5] X. Jia. Image recognition method based on deep learning, 29th Chinese Control Decis. Conf., 2017, pp. 4730–4735.

[6] A. Karpathy. Deep Visual-Semantic Alignments for Generating Image Descriptions. CVPR, 2015.

[7] M. Jaderberg, K. Simonyan, A. Vedaldi, A. Zisserman, and C. V Dec. Reading Text in the Wild with Convolutional Neural Networks. arXiv:1412.1842v1[cs.cv], 2014.

[8] A. Owens et al. Visually Indicated Sounds. arXiv:1512.08512v2[cs.cv], 2016.

[9] Hussain, Mahbub & Bird, Jordan & Faria, Diego. A Study on CNN Transfer Learning for Image Classification. UKCI ,2018

[10] R. Bakshi, "Feature Detection for Hand Hygiene Stages", Preprint ArXiv:2108.03015, 2021.

[11] M. Ivanovs, R. Kadikis, A. Elsts, M. Lulla, and A. Rutkovskis. Automated Quality Assessment of Hand washing. Pre-print ArXiv:2011.11383v2, 2020.

[12] R. Bakshi. " WHO-Hand Hygiene Gesture Classification System", Preprint ArXiv:2110.02842v1, 2021

[13] Accessed Online: Transfer learning and Fine Tuning, URL: https://keras.io/guides/transfer_learning/

[14] C. Google, "Xception: Deep Learning with Depthwise Separable Convolutions", CVPR 2017.

[15] K. He, X. Zhang, S. Ren and J. Sun. Deep Residual Learning for Image Recognition. IEEE Conference on Computer Vision and Pattern Recognition (CVPR), 2016, pp. 770-778, doi: 10.1109/CVPR.2016.90.

[16] C. Szegedy, V. Vanhoucke, J. Shlens, and Z. Wojna, "Rethinking the Inception Architecture for Computer Vision", CVPR 2016.

[17] Accessed Online: Video Classification with Keras and Deep Learning.
URL: https://www.pyimagesearch.com/2019/07/15/video-classification-with-keras-and-deep-learning/

[18] Accessed Online: Keras Applications URL: https://keras.io/api/applications/.